# Language in vivo vs. in silico: Size matters but Larger Language Models still do not comprehend language on a par with humans due to impenetrable semantic reference

Vittoria Dentella[1], Fritz Günther[2], Evelina Leivada[3,4]

[1] Department of Brain and Behavioral Sciences, University of Pavia, Pavia, Italy
[2] Institut für Psychologie, Humboldt-Universitat zu Berlin, Berlin, Germany
[3] Departament de Filologia Catalana, Universitat Autònoma de Barcelona, Barcelona, Spain
[4] Institució Catalana de Recerca i Estudis Avançats (ICREA), Barcelona, Spain

Corresponding Author:
Vittoria Dentella[1]
Email address: vittoria.dentella@unipv.it

## Abstract

Understanding the limits of language is a prerequisite for Large Language Models (LLMs) to act as theories of natural language. LLM performance in some language tasks presents both quantitative and qualitative differences from that of humans, however it remains to be determined whether such differences are amenable to model size. This work investigates the critical role of model scaling, determining whether increases in size make up for such differences between humans and models. We test three LLMs from different families (Bard, 137 billion parameters; ChatGPT-3.5, 175 billion; ChatGPT-4, 1.5 trillion) on a grammaticality judgment task featuring anaphora, center embedding, comparatives, and negative polarity. N=1,200 judgments are collected and scored for *accuracy*, *stability*, and *improvements in accuracy* upon repeated presentation of a prompt. Results of the best performing LLM, ChatGPT-4, are compared to results of n=80 humans on the same stimuli. We find that humans are overall less accurate than ChatGPT-4 (76% vs. 80% accuracy, respectively), but that this is due to ChatGPT-4 outperforming humans only in one task condition, namely on grammatical sentences. Additionally, ChatGPT-4 wavers more than humans in its answers (12.5% vs. 9.6% likelihood of an oscillating answer, respectively). Thus, while increased model size may lead to better performance, LLMs are still not sensitive to (un)grammaticality the same way as humans are. It seems possible but unlikely that scaling alone can fix this issue. We interpret these results by comparing language learning in vivo and in silico, identifying three critical differences concerning (i) the type of evidence, (ii) the poverty of the stimulus, and (iii) the occurrence of semantic hallucinations due to impenetrable linguistic reference.

## Introduction

Large Language Models (LLMs) are a dominant methodological resource in Natural Language Processing, the field of Artificial Intelligence striving to render natural language manipulable by computers. While surface similarities between natural language and text generated by LLMs are pervasive, the jury is still out with respect to whether the models' linguistic ability can be described as qualitatively and quantitatively equal to that of humans [1–4, cf. 5– 8]. LLMs are trained on contextually bound next-word prediction, a task which relies on the linear relations holding between words or their subparts. However, the relations between linguistic elements in human language are known to be of both structural and semiotic nature [9, 10]: Language is understood as a system of form-meaning associations whose joint regulation and knowledge about real-world application gives rise to linguistic behavior [7, 11–13]. While LLMs are able to learn structure and form, there exist differences in the learning regimes —i.e., some LLMs are trained solely on text, whereas others are also exposed to images and, most recently, sounds [14]—, which are coupled with the absence of sensory information, pragmatic functions, and communicative intent [15]. It is possible that such divergence in learning may give rise to some differences in the linguistic performance of humans vs. LLMs, the extent of which is not fully determined yet.

One domain where both similarities and differences have been noted concerns grammar. Humans intuitively possess awareness of the well- or ill-formedness of a sentence, something which translates into robust and replicable judgments in tasks designed to reveal what falls within people's grammar [16, 17, cf. 18–20]. Do LLMs possess the inductive biases necessary to tell apart grammatical from ungrammatical sentences? For agreement phenomena, for instance, which have been extensively studied, evidence is mixed. Some work showed that networks trained in a language-modeling setting struggled to succeed at subject-verb number agreement prediction, and that the performance of the same models explicitly trained for number agreement worsened when linear and structural information conflicted [21]. Building on this work, Gulordava et al. [22] trained a model that could formulate largely accurate number agreement predictions, even if trained solely on language modelling. Later, Mitchell and Bowers [23] retrained Gulordava et al.'s [22] model after applying transformations to their original training data for English, effectively introducing illicit structural relations. They found that, for all three modified dataset versions, the Gulordava model learned agreement patterns, as if handling licit natural structural dependencies. The influence of the properties of training corpora for tasks targeting agreement was further demonstrated in Arehalli and Linzen [24, see also 25].

While agreement patterns are central to the investigation of LLM's capability to handle hierarchical language phenomena, research has been carried out in relation to other linguistic constructs too. For instance, additional important domains of investigation concern the ability of LLMs to handle negation [26, 27], to track syntactic states (i.e., the syntactic conditions which are necessary for a sentence to license another sentence [28]), or to mechanistically develop neural units that are sensitive to the hierarchical organization of words [29]. Additionally, LLMs have become the subject of language acquisition research: able to give rise to human-like language behavior by making use of statistical means, they contribute insights into the debate over whether statistical learning suffices for the development of language [6, 7]. This, in turn, has determined an interest over the development of models that learn from a human-like quantity of input data [30, 31]. Overall, the predictions and internal representations of neural-network-based LLMs have been translated into competence over a range of morphosyntactic phenomena, allowing for a better characterization of their linguistic competence [32].

These works belong to the research area associated with the experimental analysis of the linguistic capacities of deep neural networks, called *LODNA* (i.e., linguistically-oriented deep net analysis [33]), which aims at probing the linguistic knowledge encoded in networks. In this line of research, the language capabilities of LLMs are often evaluated through obtaining direct measurements of the probabilities they assign over minimal pairs of sentences, one of which is grammatical and the other one ungrammatical [34]. The expectation is that the LLM assigns a lower probability to the ungrammatical prompt of the minimal pair. For example, the models in Wilcox et al. [35] assign higher surprisal values (namely, "the extent to which [a] word or sentence is unexpected under the language model's probability distribution"; p. 212) to illicit filler-gap constructions as opposed to grammatical ones. Overall, this method has been linked to claims that the models have internalized a notion of grammaticality, as evidenced through their good performance in probability assignment [34, 36].

While obtaining probabilities from LLMs gives rise to valuable results, this is not a language task [37, see also 36]. Probabilities amount to numbers that *humans* interpret in relation to certain linguistic dimensions such as grammaticality, semantic plausibility, or pragmatic coherence. Put another way, when we elicit probabilities assigned to strings of words from a model, we obtain values that we need to translate into claims about language, but we do *not* obtain linguistic behavior.

An alternative method of LLM evaluation involves prompting, that is, asking the model to provide a language output (e.g., a judgment of well-formedness) as a response to a given prompt, based on whether the latter complies with or deviates from the model's next-word predictions [38]. Prompting brings forward the behavioral interpretability of the models; and it is a useful evaluation method that can inform both linguistic theory and our understanding of deep neural models [39]. Prompting in LLMs follows a decades-long tradition of running such experiments with humans [40]. Though not exempt from critiques [34, 36], the prompting method has the potential to reveal possible similarities and differences between humans and LLMs through obtaining and analyzing language outputs (see [37, 41] on why probabilities are not a good index of grammaticality and [36] for counterarguments).

While for image-generation models there already exists evidence that they face challenges with compositionality [42], interpretative constraints [43], and common syntactic processes more broadly [44], for the LLMs' ability to recognize (un)grammaticality and impossible language, evidence is still rather scarce [45]. Recently, Dentella et al. [46] carried out a systematic assessment of the performance of three LLMs on a grammaticality judgment task and subsequently checked it against the performance of humans on the same stimuli. They found that while humans are aware of grammatical violations even for hard-to-parse sentences, the LLMs struggled in providing consistent, accurate judgments especially for ungrammatical sentences, marking a stark difference from human performance.

The present work investigates whether model scaling mitigates such differences. More specifically, we ask whether the parameter size of LLMs affects their accuracy and response stability in grammaticality judgment tasks. Consequently, we employ the number of training parameters as the only predictor of LLM's performance. As scaling could be associated with improved performance in the linguistic domain, our analysis can shed light on whether viewing LLMs as cognitive theories that make accurate predictions about language [6, 7] is supported by the potentially better performance of bigger models. If LLMs are to be treated as theories of language [33], the ability to discern whether a prompt falls within the range of predicted possible outputs or not is a prerequisite; and the role of model size in this context is important to determine.

To this end, we compare three LLMs on a grammaticality judgment task. We ask whether scaling in terms of numbers of parameters substantially improves accuracy (RQ1), stability (i.e., providing the same answer when a prompt is repeated) (RQ2), and whether accuracy and/or stability improve when a prompt is presented multiple times (RQ3). Lastly, we compare the results of the best performing LLM, ChatGPT-4, to those of n=80 human subjects (reported in [46]).

## Materials & Methods

The present work tests four grammatical phenomena, using tasks that have already been validated in previous research: anaphora [47]; center embedding [48]; comparative sentences [49]; and negative polarity items [50]. These phenomena are chosen because they posed significant challenges for some LLMs; specifically, GPT-3/text-davinci-002, GPT-3/text-davinci-003, and ChatGPT-3.5 [46]. For each phenomenon, 10 sentences (which are split for condition: 5 grammatical, 5 ungrammatical) are tested. Each sentence is given 10 times to each LLM through the elicitation question "Is the following sentence grammatically correct in English? ___". All prompts are merged in a unified pool, randomized, and subsequently administered in random order. Yes/no binary judgments are obtained which are coded for accuracy (1 for accurate answers, 0 for inaccurate answers) and stability (1 for change from the previous judgment given to the same prompt, 0 for absence of change). For each of the three tested LLMs, 400 judgments are collected, resulting in n=1,200 total judgments. Table 1 presents two sample sentences per phenomenon. The full list of prompts can be found at https://osf.io/m6yda/?view_only=135e7be5a131458e89acc05c74247b09.

| **Anaphora** |
|---|
| Grammatical (matching anaphoric reference), ungrammatical (mismatching anaphoric reference)<br><br>1. The police officer who aided the brilliant detectives reportedly disguised himself to get more information<br>2. *The paranoid foreman who supervised the coal miners presumably saved themselves after the awful cave-in |
| **Center embedding** |
| Grammatical (all verbs present), ungrammatical (one verb missing)<br><br>1. The picture that the artist who the school had expelled for cheating was hurriedly copying was printed in a magazine<br>2. *The shirt that the seamstress who the immigration officer had investigated last week needed to be dry cleaned |
| **Comparative sentences** |
| Grammatical (plural comparison set), ungrammatical (singular comparison term)<br><br>1. This month more one-year-olds took their first steps than 10-month-olds did<br>2. *Last month more couples had their second child than Sandra's sister did |
| **Negative polarity items** |
| Grammatical (negation licensor), ungrammatical (no negation licensor)<br><br>1. No customers that the salesmen assisted have expressed any optimism for a full refund<br>2. *The lawyers that no businessmen respected have ever received criticism for a bad trial |

Table 1: Sample sentences (one grammatical, one ungrammatical) for each of the four tested phenomena.

The grammaticality judgment tasks were administered to three LLMs set on default interface parameters: Bard [51], ChatGPT-3.5 [52], and ChatGPT-4 [53]. The results by Bard and ChatGPT-4 were collected in September 2023, and are available at https://osf.io/m6yda/?view_only=135e7be5a131458e89acc05c74247b09. They are compared to previous results obtained from ChatGPT-3.5 in February 2023 and from n=80 humans (both datasets presented in [46]), using the same tasks. Bard and ChatGPT-4 were tested after ChatGPT-3.5's results were made public in the Open Science Framework repository of Dentella et al. [46]. It is therefore possible that Bard and ChatGPT-4 have been exposed to the testing materials [41]: while such exposure is bound to affect the models' performance, it should also place the models –which all include Reinforcement Learning from Human Feedback (RLHF) in their training– in a better position to perform aptly in the tasks.

## Motivation

Testing LLMs at their user interfaces presents with two benefits. First, it obtains language outputs as results, so that LLMs can be evaluated on their default capacity to manipulate language in a way that allows meaningful interaction with humans. While methods such as probability readings also offer insight into a model's capacities [34], these methods obtain numeric outputs which are not natural language [37, 41]. Second, prompting LLMs with grammaticality judgment tasks has the potential to inform the field of linguistics. These tasks represent an established methodology which has been used in research with humans participants for decades [17, cf. 18], and the use of the same elicitation method for LLMs is motivated by the need to compare the two agents, models and humans, in order to determine the similarities and differences that exist between the two in processing language. Last, probability distributions are not available for several existing models, including ChatGPT-4 [53], and this limitation suggests the need of searching for alternative ways of evaluation.

**Reproducibility and Model Description**

The LLMs were tested directly at the interface, and the code employed for the analysis of the obtained results is available at https://osf.io/m6yda/?view_only=135e7be5a131458e89acc05c74247b09.

Bard is a LaMDA-based conversational model (137 billion parameters) pre-trained on 1.5 trillion words [51], ChatGPT-3.5 (175 billion parameters) is a transformer-based model fine-tuned from a GPT-3.5 model [52], and ChatGPT-4 is a 1.5 trillion parameter multimodal model [53]. All three models include RLHF in their training. These LLMs were chosen as state-of-the-art at the time of testing. Additionally, the choice of models deployed at mass-scale and featuring RLHF was guided by an interest for evaluating models which, in addition to being pre-trained on the task of language modelling, benefitted from further fine-tuning and human intervention so as to maximize utility, which was hypothesized to lead to better performance compared to pre-trained only, niche LLMs. Lastly, the substantial difference in number of parameters between Bard and the two ChatGPT models, together with the difference between ChatGPT-3.5 and ChatGPT-4, allowed the evaluation of scaling effects both across and within model families. As the models were not subject to training, no computing infrastructure was necessary; all models were tested at their respective commercial interfaces using a personal computer running on Windows 11 Pro. Data preprocessing does not apply.

## Results

Overall, we have four different responding *agents*: Three different LLMs that will be referred to as *models*, and humans. In the first analyses, we focus on the models. The data were analyzed using (Generalized) Linear Mixed-Effect Models [54]. All (G)LMMs included random intercepts for items (sentences) nested within the four different phenomena. Starting from an intercept-only (G)LMM, we used likelihood-ratio tests comparing a (G)LMM containing a given parameter to a model without it in order to test for particular effects, thereby identifying the optimal (G)LMM structure. For all (G)LMMs, ChatGPT-3.5 and grammatical sentences served as baseline for *model* (i.e., Bard, ChatGPT-3.5, ChatGPT-4) and *condition* (grammatical, ungrammatical), respectively. The code employed for the analyses is available at https://osf.io/m6yda/?view_only=135e7be5a131458e89acc05c74247b09.

**Accuracy**

The GLMM predicting accuracy (the likelihood of a correct response; accuracy ~ 1 + (1 | phenomenon/sentence)) was significantly improved by first adding a parameter for *model* (accuracy ~ model + (1 | phenomenon/sentence); $\chi^2(2) = 104.40$, $p < .001$), then a parameter for *condition* (accuracy ~ model + condition + (1 | phenomenon/sentence); $\chi^2(1) = 35.74$, $p < .001$), but not by an additional interaction between the two (accuracy ~ model * condition + (1 |phenomenon/sentence); $\chi^2(2) = 4.42$, $p = .110$). As can be seen in Figure 1 (dashed lines), this is due to the fact that ChatGPT-4 outperforms the other LLMs across both conditions ($\beta = 1.72$, $z = 9.11$, $p < .001$), with no significant difference between ChatGPT-3.5 and Bard ($\beta = 0.27$, $z = 1.59$, $p = .111$). All three LLMs provide fewer accurate answers for ungrammatical than grammatical sentences ($\beta = -2.20$, $z = -7.45$, $p < .001$). Importantly, ChatGPT-4 (a) reaches a very high albeit not perfect level of accuracy for grammatical sentences (93.5 %), and (b) is the only LLM that performs *above* chance level for ungrammatical sentences, thus not displaying the yes-response bias found for all other LLMs tested in this task [46]. These results suggest that an exponentially higher number of training parameters significantly improves performance, but to some extent, this remains insufficient for leveling out the discrepancy in accuracy rates between grammatical and ungrammatical sentences.

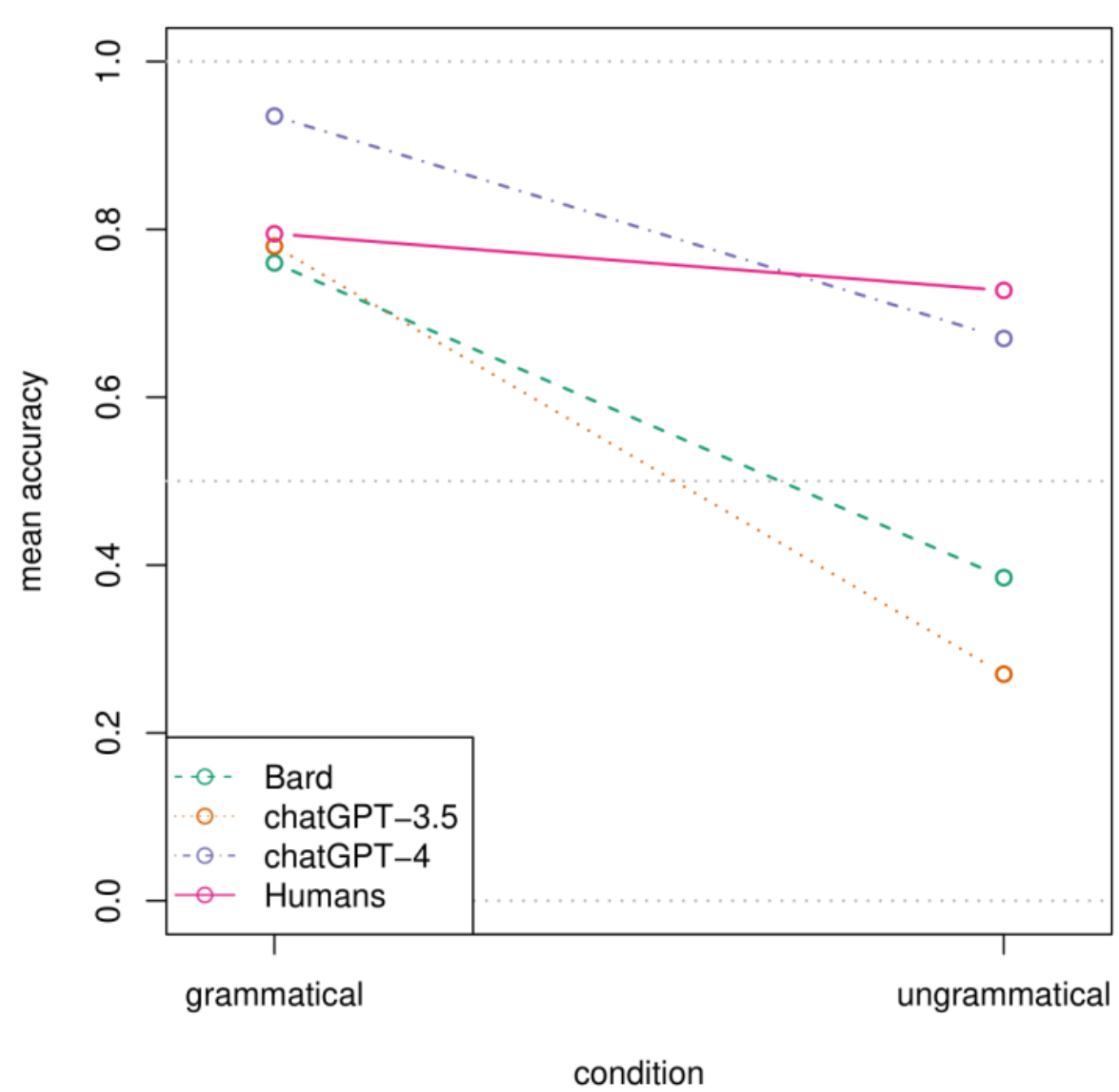

Figure 1: Mean accuracy by responding agent (i.e., Bard, ChatGPT-3.5, ChatGPT-4, and humans) and condition: All individual responses.

**Stability**

We used the same two variables to operationalize response stability employed in Dentella et al. [46]: *Oscillations*, a local trial-level measure, encode whether the response in a given trial was identical to the last response for the same sentence (thus ranging between 0 and 9 per sentence). *Deviations*, a global sentence-level measure, encode the frequency of the less frequent response (i.e., the non-preferred response) per sentence (thus ranging between 0 and 5 per sentence).

The GLMM predicting the likelihood of an oscillation was significantly improved by first adding a parameter for *model* (oscillation ~ model + (1 | phenomenon/sentence); $\chi^2(2) = 24.69$, $p < .001$), then a parameter for *condition* (oscillation ~ model + condition + (1 | phenomenon/sentence); $\chi^2(1) = 11.89$, $p < .001$), but again not by an additional interaction between the two (oscillation ~ model * condition + (1 | phenomenon/sentence); $\chi^2(2) = 5.29$, $p = .071$). Again, this reflects that ChatGPT-4 provides more stable/less oscillating responses than the other LLMs across both conditions ($\beta = -0.92$, $z = -4.48$, $p < .001$), with no significant difference between ChatGPT-3.5 and Bard ($\beta = -0.08$, $z = -0.45$, $p = .650$; see Figure 2, A, dashed lines). As in the case of accuracy, for stability too there is a difference in condition, as all three LLMs provide less stable/more oscillating answers for ungrammatical than grammatical sentences ($\beta = 0.88$, $z = 3.66$, $p < .001$).

The GLMM for the number of deviations was significantly improved by first adding a parameter for model (deviation ~ model + (1 | phenomenon/sentence); $\chi^2(2) = 11.60$, p < .001), but not by an additional parameter for condition (deviation ~ model + condition + (1 | phenomenon/sentence); $\chi^2(1) = 1.87$, p = .172), or this condition parameter plus an interaction between the two (deviation ~ model * condition + (1 | phenomenon/sentence); $\chi^2(3) = 2.60$, p = .458). Again, we see that ChatGPT-4 provides more stable responses (fewer deviations; b = -1.10, t(114) = -3.03, p = .003; see Figure 2, B, dashed lines), with no difference between ChatGPT-3.5 and Bard (b = -0.025, t(114) = -0.07, p = .945).

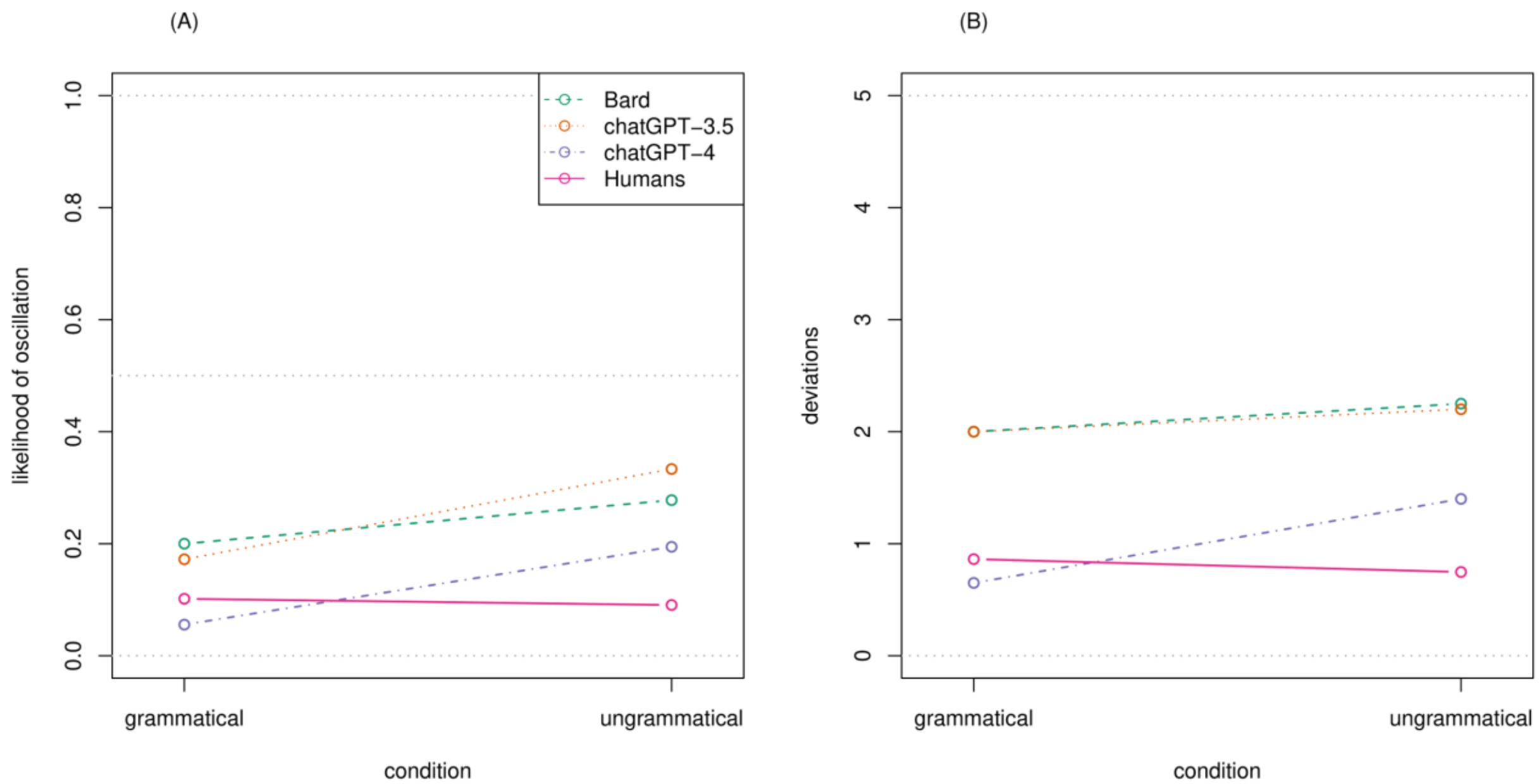

Figure 2: Response instability by responding agent (i.e., Bard, ChatGPT-3.5, ChatGPT-4, and humans) and condition, as measured by (A) the likelihood of oscillations, and (B) the number of deviations per sentence.

**Interplay between Stability and Accuracy**

A lack of stability puts a strict upper limit on accuracy: if there are deviations, the maximum possible accuracy rate is reached if all non-deviating (i.e., preferred) responses are correct. To test whether the LLMs are accurate in their majority answers, we performed the accuracy analysis only on the preferred answer per sentence. The GLMM predicting this accuracy was significantly improved by first adding a parameter for model (bin_cor ~ model + (1 | phenomenon/sentence); $\chi^2(2) = 10.51$, $p = .001$), then a parameter for condition (bin_cor ~ model + condition + (1 | phenomenon/sentence); $\chi^2(1) = 47.16$, $p < .001$), and again no interaction between the two (bin_cor ~ model * condition + (1 | phenomenon/sentence); $\chi^2(2) = 2.96$, $p = .229$). This repeats the pattern for the individual trial-level responses, only in a more pronounced manner (more correct responses for grammatical sentences and fewer for ungrammatical ones), especially for Bard and ChatGPT-3.5 (see Figure 3, dashed lines).

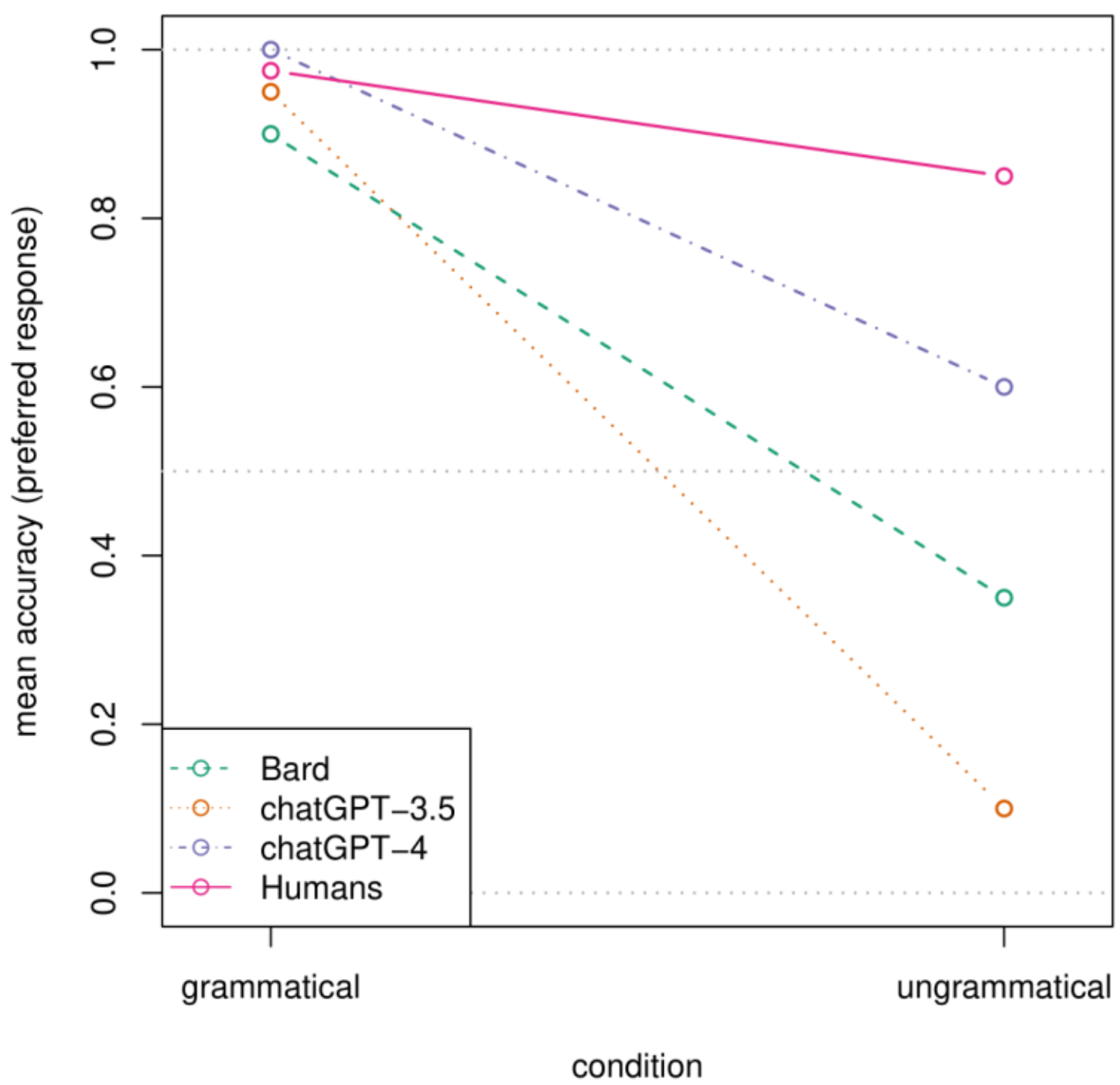

Figure 3: Mean accuracy by responding agent (i.e., Bard, ChatGPT-3.5, ChatGPT-4, and humans) and condition: Preferred (i.e., more frequent) responses per sentence.

**Effects of Repetitions**

Since the LLMs are not perfectly stable in their responses, it might be possible that they improve over repeated presentations of the same sentence and become more accurate or more stable.

Indeed, we find that the GLMM for accuracy including the interaction between model and condition is further improved by adding a parameter for repetitions (encoding how often the same sentence had been presented; accuracy ~ model * condition + repetition + (1 | phenomenon/sentence); $\chi^2(1) = 4.88$, p .027), and even further by adding a three-way interaction between model, condition, and repetitions (as well as all required lower-level interactions; accuracy ~ model * condition * repetition + (1 | phenomenon/sentence); $\chi^2(5) = 22.15$, $p < .001$). As can be seen in Figure 4 (dashed lines), this captures that some models tend to improve for some conditions over repetitions (ChatGPT-4 for grammatical sentences, or Bard for ungrammatical sentences), while others even decline in accuracy (ChatGPT-3.5 for both conditions, and Bard for grammatical sentences). These conflicting tendencies show that performance improvement over repetitions does not necessarily correlate with model size as Bard, the model with the least number of training parameters, shows improvement in a condition where ChatGPT-4 does not and where ChatGPT-3.5 worsens.

When it comes to stability in the form of its trial-level measure, oscillations, we do not find that the addition of any repetition effect –neither main effect nor any two- or three-way interaction– improved the GLMM reported in the *Stability* section (with its main effects of *model* and *condition*); all *p*s > .075. We thus find no evidence that response stability changes over repeated presentations of the same sentence, with model size playing no role in this respect.

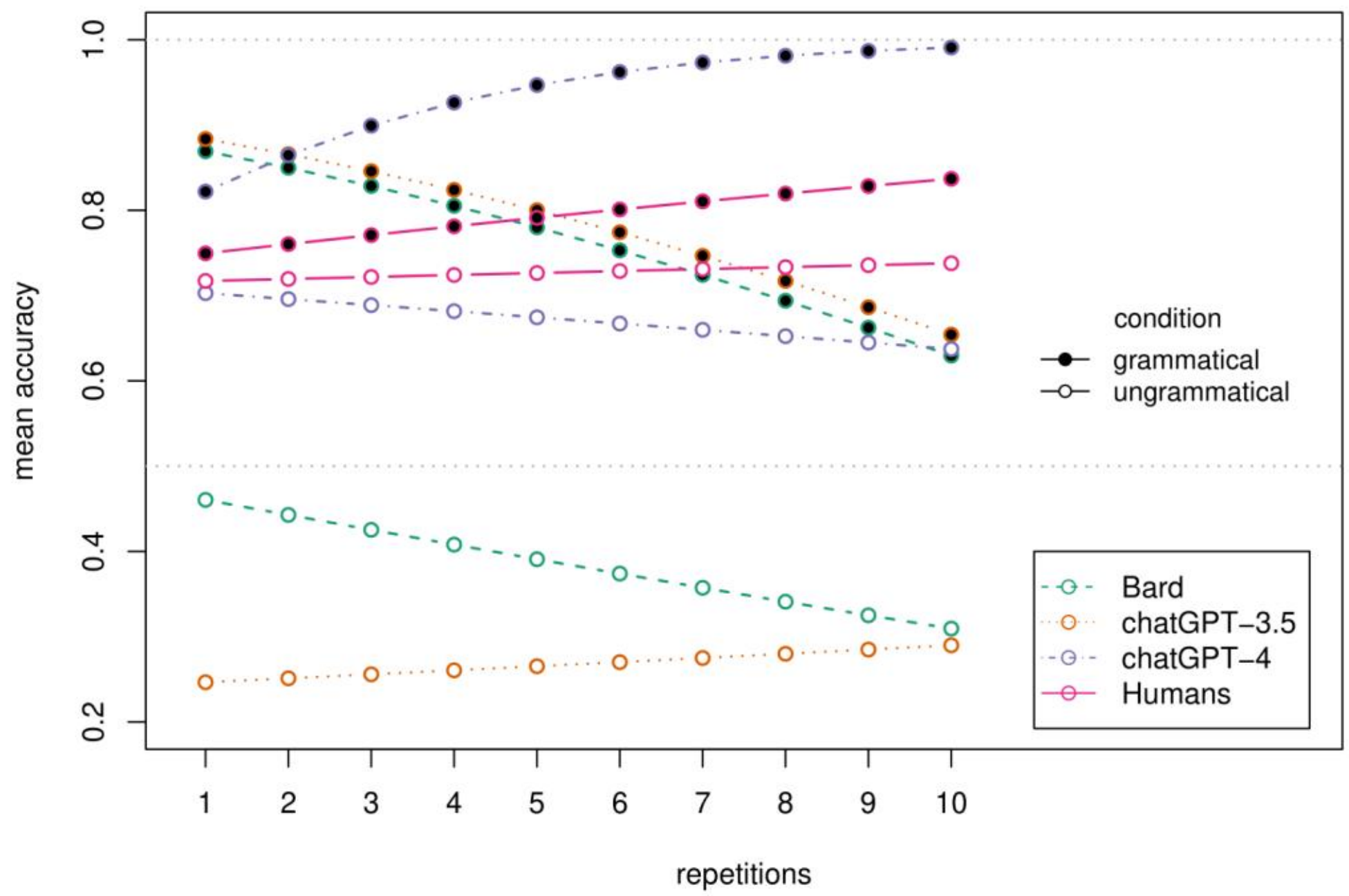

Figure 4: Mean accuracy by number of repetitions, condition, and responding agent (i.e., Bard, ChatGPT-3.5, ChatGPT-4, and humans) (the points indicate GLMM predictions rather than observed data).

**Comparison with Human Data**

Dentella et al. [46] found that LLMs were overall less accurate and stable than humans in this task. Here, we examine if this pattern still holds for ChatGPT-4, the best-performing LLM. To this end, we estimated the same (G)LLMs as previously, only replacing the model parameter with *type* (a two-level factor comparing the reference condition ChatGPT-4 to humans).

For the accuracy data (Figure 1, pink vs. violet lines), we find an interaction between *type* (i.e., human vs. ChatGPT-4) and *condition* (accuracy ~ type * condition + (1 | phenomenon/sentence); $\chi^2(1) = 14.84$, $p < .001$), to the effect that ChatGPT-4 provides more correct responses than humans for grammatical sentences, but fewer for ungrammatical ones. Averaging over both conditions, humans produce fewer correct responses (76.1%) than ChatGPT-4 (80.3%; $\beta = -0.53$, $z = -3.77$, $p < .001$). This marks an important difference from the results of Dentella et al. [46], who did not find a significant main effect of type when comparing ChatGPT-3.5 with humans, strongly suggesting that model size indeed matters. For the accuracy of preferred responses (i.e., non-deviating responses; Figure 3, pink vs. violet lines), we found no interaction (bin_cor ~ type * condition + (1 | phenomenon/sentence); $\chi^2(1) = 1.97$, $p = .161$) and no main effect of type (accuracy ~ type + condition + (1 | phenomenon/sentence); $\chi^2(1) = 3.31$ $p = .069$), indicating no difference between humans and ChatGPT-4 for this outcome variable.

Turning to response stability, for oscillations (Figure 2, A, pink vs. violet lines) we again find an interaction between *type* and *condition* (oscillation ~ type * condition + (1 | phenomenon/sentence); $\chi^2(1) = 16.63$, $p < .001$), again to the effect that ChatGPT-4 oscillates less for grammatical sentences but more for ungrammatical ones. Overall, the likelihood of an oscillation is lower for humans (9.6%) than for ChatGPT-4 (12.5%; $\beta = -0.38$, $z = -2.19$, $p = .029$). The same interaction pattern emerges for deviations (Figure 2, B, pink vs. violet lines; deviation ~ type * condition + (1 | phenomenon/sentence); $\chi^2(1) = 4.44$, $p = .035$), but without a significant overall difference in the number of deviations between humans (0.81) and ChatGPT-4 (1.03; $b = -0.29$, $t(766) = -1.41$, $p = .159$). This picture largely corresponds to that of Dentella et al. [46], indicating that while size matters, it is not sufficient to annihilate all the differences between the linguistic performance of humans and LLMs across fronts and measures.

When analyzing changes in accuracy over repetitions, we also find a significant three-way interaction between *type*, *condition*, and *repetition* (accuracy ~ type * condition * repetition + (1 | phenomenon/sentence); $\chi^2(1) = 7.02$, $p = .008$ when compared to a model with all two-way interactions). As can be seen in Figure 4 (pink vs. violet lines), for grammatical sentences ChatGPT-4's target performance increases ($\beta = 0.376$, $p = .005$ for the *repetition* parameter) more than that of humans ($\beta = -0.306$, $p = .022$ for the interaction between *type* and *repetition*). For ungrammatical sentences, there is no statistically significant difference between the repetition effect for ChatGPT-4 and humans ($\beta = -0.04$, $p = .514$, for the effect of *repetition* and $\beta = 0.05$, $p = .382$, for the interaction between *type* and *repetition* when setting "ungrammatical" as the reference category for *condition*).

In summary, while ChatGPT-4 overall achieves higher accuracy than humans, this is largely due to its performance in grammatical sentences. For ungrammatical sentences, it is less accurate (with this accuracy decreasing over repetitions), and its responses are less stable.

## Discussion

In this work we compare three LLMs (Bard, ChatGPT-3.5 and ChatGPT-4) on a grammaticality judgement task featuring four linguistic phenomena. We ask whether an increased number of training parameters translates into better performance which, for the task at hand, consists in providing accurate grammaticality judgments (RQ1), that are consistent across repetitions of the same prompt (RQ2) or, if they vary, they nonetheless converge towards accurate or stable responses over repetitions of the same prompt (RQ3). The performance of Bard and ChatGPT-3.5, respectively the smallest and the second-smallest model in the analyses, is comparable on all measures with one exception: When repeatedly exposed to the same stimuli, Bard's responses on grammatical sentences tend to become accurate, whereas the performance of ChatGPT-3.5 worsens for both conditions. Additionally, the comparable performance of these two models implies that ChatGPT-3.5 fails to outperform Bard, something which provides counterevidence to scaling being invariably associated with improvements in model performance. On the other hand, ChatGPT-4, the largest tested model, significantly outperforms both Bard and ChatGPT-3.5. As opposed to Bard, however, it fails to show increases in accuracy upon repeated exposure to ungrammatical stimuli. Table 2 provides a summary of the key findings.

| | |
|---|---|
| **RQ1. Accuracy** | |
| 1. | All LLMs are more accurate on grammatical than on ungrammatical sentences |
| 2. | Bard and ChatGPT-3.5 perform comparably |
| 3. | ChatGPT-4 is the only LLM that performs above chance level for ungrammatical sentences |
| 4. | ChatGPT-4 is more accurate than humans on grammatical sentences, but less accurate on ungrammatical ones |
| **RQ2. Stability** | |
| 1. | All LLMs are more stable on grammatical than on ungrammatical sentences |
| 2. | Bard and ChatGPT-3.5 perform comparably |
| 3. | ChatGPT-4 outperforms the other LLMs |
| 4. | ChatGPT-4 wavers in responses more than humans (for oscillation measures) |
| **RQ3. Effects of Repetitions** | |
| 1. | Bard improves in accuracy for ungrammatical sentences, but declines for grammatical ones |
| 2. | ChatGPT-3.5 declines in accuracy for both grammatical and ungrammatical sentences |
| 3. | ChatGPT-4 improves in accuracy for grammatical sentences |
| 4. | ChatGPT-4's improves in accuracy for grammatical sentences only, while humans improve across conditions |

Table 2: Summary of key findings per Research Question (RQ).

Taken together, this evidence shows that, while the discrepancy between ChatGPT-4 and the other two LLMs suggests that increases in size correlate with better performance, this relationship is not strictly linear. Particularly, our results show that even the best-performing model in our analyses, ChatGPT-4, does not behave comparably to humans: ChatGPT-4's accuracy rate in grammatical sentences surpasses that of human subjects, but it is lower than that of humans on ungrammatical stimuli, where it further decreases upon repeated exposure. Also, the model is overall less stable, again especially for ungrammatical sentences. Human responses are instead largely accurate across conditions, stable, and their accuracy increases with repetitions across *both* conditions, with such properties being present all at the same time.

LLMs are applications conceived with the end goal of emulating human linguistic behavior. In this context, linguistic tasks can be employed to determine whether LLMs have indeed mastered human language. Acceptability judgments in humans represent an established methodology for determining what forms part of a person's internalized grammar [40], and such judgments in humans are both robust and replicable [16, 17]. The fact that ChatGPT-4, the best-performing model in our testing, does not identify syntactic anomalies at ceiling across conditions —neither the absolute ceiling of perfect accuracy nor the baseline set by human speakers— is a sign that the structural generalizations behind the grammatical well-formedness of grammatical prompts are not encoded in the model.

While humans are subject to cognitive constraints (e.g., working memory limitations, distraction, fatigue, idiolectal preferences that give rise to interspeaker variation, etc.) that can cause occasional failure in providing target acceptability judgments, this performance does not necessarily reflect our competence [55]. In other words, performance errors in non-pathological subjects have roots in and can be explained by appealing to shallow processing and heuristics of cognition [56, 57]. On the other hand, in absence of a clear theory of their cognitive abilities and constraints, this is not the case for LLMs. In this sense, language performance in humans presupposes and relies on competence, but this relation may not hold for LLMs: They have impressive abilities for generating human-like text, amounting to an almost impeccable performance, but their competence is still a gray zone. Alternatively, if one wants to argue that their competence is also impeccable and human-like, it remains to be explained what exactly makes them less stable in their judgments upon repeated prompting.

Recently, LLMs have been claimed to have mastered natural language and, consequently, to be able to act as cognitive theories [6, 58]. For LLMs to be compared to humans, however, linguistic behavior that is qualitatively on a par with that of humans is a necessary condition of adequacy [2]. Furthermore, this condition should be met given exposure to (or training on) the same amount of linguistic data. As opposed to LLMs, humans do not need exposure to gigantic datasets to acquire a language [59]. Notwithstanding training on datasets that virtually span the whole internet, however, the models we tested still cannot identify grammatical errors in the stimuli in a stable way. In this respect, a second point of departure between LLMs and humans concerns the nature (in addition to the amount) of linguistic data employed for learning. All the LLMs tested here benefit from massive amounts of human intervention in the form of RLFH, which contributes both the hard-engineering of target responses, and the introduction of *negative evidence* in training. This means that LLMs are explicitly instructed by humans on what is *not* grammatical in a language.

For humans, limited exposure to data is made up for by our innate endowment for language, as evidenced in the creolization of pidgins, or the spontaneous development of sign languages such as Al-Sayyid Bedouin Sign Language. To attempt a direct comparison, the innate endowment of LLMs consists of their architecture (e.g., long short-term memory, convolutional or transformer networks) which, input data being equal, is responsible for different behaviors. During the learning phase, a LLM picks up patterns in the data and the learned amount translates into parameters: The higher the number of parameters a model has, the more the accumulated knowledge that can be deployed for a task. The LLMs discussed in the present work have 137 and 175 billion parameters (Bard and ChatGPT-3.5, respectively) and 1.5 trillion parameters for ChatGPT-4. To quantify, ChatGPT-4's size in this respect is approximately 10.9 and 8.6 times larger than Bard and ChatGPT-3.5's respectively (for comparison, in the OpenAI model family GPT-2 had 1.5 billion parameters and GPT-1 had 117 million parameters).

While the massive upgrade in scale has significantly reduced quantitative differences in the performance between humans and ChatGPT-4, taken here as representative of state-of-the-art language modelling, these differences persist to a lesser degree, not so much in terms of accuracy but certainly in terms of stability. While ChatGPT-4 surpasses humans in accuracy for the grammatical condition, it underperforms in the ungrammatical condition and, more importantly, it fluctuates in its responses, further failing to demonstrate improvement with repetitions. These facts are difficult to reconcile with the proposed idea that LLMs run counter to claims of language innateness in humans [6], which rest on the ground that a LLM able to reproduce language is proof that language can be learned without the need for an innate endowment for it. While our results show that LLMs fail to fully reproduce human linguistic behavior, this is neither evidence for innateness in humans, nor support for any specific theory of innateness. In this sense, our interpretation of the results agrees with Cuskley et al.'s [13] claim that LLMs tell us little about human language development and evolution. We also note that while our results suggest that the best available model to date still fails to identify sentence errors on a par with humans, other methods of prompting that alter the testing regime (e.g., by restricting rereading in [8]) may provide different results, minimizing or amplifying the differences between humans and models. This variation is expected as there are testing environments that would make the models outperform humans, for instance, by manipulating memory constraints that are obviously not uniform in humans vs. models. From this perspective our findings do not suggest that LLMs will always outperform humans across all language tasks, neither do they imply that LLMs are principally unable to learn language, but rather that they do not possess its mastery at their current stage of development. Scaling indeed matters, as the largest model tested here performs better than the smaller models tested in previous work [46]; however, alternative explanations for this better performance are possible (e.g., algorithmic fudging after exposure to tasks and datasets that have been made public [41]).

An additional issue worth considering concerns the LLM (in)capability to encode grammaticality. Indeed, it is not obvious that LLMs should be expected to represent a notion of the grammatical well- or ill-formedness status for the sentence material they produce or receive as input [60], unless one believes that models are human-like in terms of language generalization capabilities [36]. If they are human-like, they are expected to perform like humans in language tasks that work for humans. While the results of the present work suggest that this expectation is not fully borne out, it has been recently argued that prompting, as opposed to probability measurements, might not be the most suitable method to provide conclusive evidence as per whether a model possesses a given linguistic generalization [34, 36]. Probability measurements, however, do not necessarily determine the boundaries of an internalized grammar [37, 41]. In other words, a model's judgments are based on likelihood, which does not necessarily reflect grammaticality [2, see also 20]). If a model's capacity to encode grammaticality is assessed on the basis of probability measurements [e.g., 21, 22, 29, 34], whereby higher probability coincides with grammaticality and lower probability with ungrammaticality, it is not straightforward that the model *recognizes* the lower-probability form as grammatically incorrect. It is humans who do the mapping between lower-probability and ungrammaticality and arrive to this interpretation, not models.

Do LLMs comprehend language on a par with humans? To answer this question, it is important to take into account that the methodology one uses can influence the results. When the aim is to determine whether LLMs possess *human-like* language capacities, any task that is suitable for humans should in principle be applicable to LLMs, leaving physiological constraints aside. In other words, the argument that prompting is not a suitable method for testing LLMs [34, 36] is not strong because it runs into a paradox [41], whereby LLMs require methodologies that are *not* applicable to humans (e.g., we cannot peek into people's neurons and see what probability they assign to a string of words) in order to be evaluated as having achieved human-like abilities. This is the 'human-like paradox': the models are simultaneously both human-like (in terms of how well their probabilities align with human judgments) and not human-like (in the sense that certain tasks and methods that work well for humans are deemed as inappropriate for them).

Overall, the results obtained through different methodologies (i.e., grammaticality judgment prompting vs. probability measurements) pose issues of comparability, which need to be collectively tackled before any firm conclusions can be drawn. A similar reasoning can be applied to prompting styles: While humans are sensitive to acceptability judgment task features such as the single vs. joint presentation of sentences with contrasting grammaticality status [61], human judgments are consistent at both the individual and at the group level *regardless* of task features. For LLMs, instead, different wordings employed for a prompt can contribute substantial differences in the obtained results within and across LLMs [62]. While it is true that attempting several prompting styles can aid the LLMs in providing correct answers, it is not clear that such practice would contribute to establishing a fair comparison with humans (cf. [63] for considerations on the comparability of human/LLM evaluation methods).

In this context, perhaps the most important question is whether scaling can bridge differences between the language abilities of humans vs. models. This discussion is linked to debates about human uniqueness: While the communication systems of other species bear similarities with human language, the latter is rendered unique by hallmark characteristics that, taken together, synthesize the ability to (re)combine finite sets of elements so as to create an infinite number of outputs that refer to some perception of real-world reality [64]. Among these characteristics are reference (i.e., the ability to use lexicalized concepts to refer to persons, objects, and events), compositionality (i.e., the ability to combine the meaning of the parts into a meaning of the whole, reflecting the way parts are combined), hierarchical grammatical dependencies (i.e., the ability to combine parts into a single, composite, hierarchically structured whole), duality of patterning (i.e., the ability to form discrete, meaningful units from discrete, non-meaningful units), and semanticity (i.e., the ability to develop fixed associations between specific linguistic forms and their denotation in the world) [55, 65, 66]. Bringing models into the discussion of human uniqueness, we argue that the observed differences in task performance are the consequence of a different way of 'learning'. More specifically, we argue that form training *in silico* departs from language learning *in vivo* in at least three critical ways.

The first difference concerns the type of evidence that is available to LLMs vs. humans. Specifically, while humans have access to positive evidence only [67, 68], LLMs also have access to negative evidence. For instance, GPT-family models explicitly acknowledge upon being asked about their training that "during my training process, I was exposed to a diverse dataset that includes both grammatically correct and grammatically incorrect sentences. The dataset is carefully curated to cover a wide range of language patterns, including common mistakes that humans make when writing or speaking in English. By being exposed to examples of both correct and incorrect sentences, I learn to recognize the differences and understand the rules of grammar more effectively" (response obtained by ChatGPT-3.5, July 2023). This grants LLMs with one more source of instruction than humans and allegedly better equips them for the task of identifying ungrammatical sentences. Yet, some of the LLMs tested here largely fail at this task. Assuming that LLMs possess knowledge of grammatical rules, their failure to make use of explicit instructions as per what counts as ungrammatical in a language is hard to account for [69], raising the question of whether LLMs truly possess the ability to *understand* such instructions. If LLMs are incapable of figuring out the boundaries of a language despite exposure to the relevant rules, it is unclear what type of information would suffice for them to behave in a human-like way in this domain.

A second difference boils down to the quantity of evidence available to LLMs vs. humans. By virtue of the innate properties presented above, humans can create language ex novo. An instance of this ability are pidgins and creoles, languages which emerge in multilingual societies in the absence of a common language for communication [70]; a process demonstrating that the amount of available evidence can be marginal when developing a natural language grammar. On the other hand, LLMs require scaling and vast amounts of data which, while contributing ameliorated performance through the leverage of data artifacts [71], do not bridge the *qualitative* gap with natural language [15, 41].

Third, there is the issue of impenetrable linguistic reference, that is, the struggle of text-based computational models to induce meaning from form alone [15]. While it has been argued that the natural histories of words may suffice for them to refer to the real world, and thus to carry meaning [72], the question of whether such words mean something *for* the models which produce them —as opposed to the humans who interpret LLMs' text strings— is debated [73]. Humans learn language through forming hypotheses about the input [74]. Unlike humans, LLMs only perform data-based predictions, lacking theory formation [15, 75]. The consequence is that LLMs give rise to hallucinating outputs that additionally deviate from answers neurotypical humans would provide. For instance, LLMs often fail to identify grammatically correct and semantically coherent inputs as such [3], in addition to generating outputs pertaining to non-target semantic frames [76]. In other words, it is possible that LLMs generate sequences of words which correctly pattern together, thanks to their good next-word predictions, however, the words themselves are semantically impenetrable black boxes to the models [77]. This inability to understand language [15] translates into an impossibility of learning it in a human-like sense.

Differences in (i) the quality of evidence available to LLMs, (ii) the quantity of such evidence, and (iii) the acquisition of meaning, are relevant to debates over issues of human language learning, to which LLMs have recently substantially contributed [31]. Particularly, while it is recognized that the language learning process in humans entails a statistical component [78] which can be in principle reproduced in LLMs, it is not clear that a model capable of extracting statistical regularities from its training data is likewise apt for deploying the acquired knowledge towards the development of a language system. While LLMs have been argued to set the debate over Poverty of the Stimulus arguments [59] to rest in favor of empiricist accounts of acquisition [6], the results hereby presented challenge this view: the inability of the tested LLMs to adhere to a grammaticality judgment task's demands suggests that LLMs lack the internal mechanisms that allow humans to naturally tell grammatical and ungrammatical stimuli apart [79].

In addition to these three foundational differences in language learning between humans and LLMs, one last point which merits consideration is the proprietary nature of the LLMs here tested. In addition to possible differences in their respective training data, Bard, ChatGPT-3.5 and ChatGPT-4 parameters are subject to constant changes in response to user inputs using RLHF; therefore, it is unclear whether their results are due to improved next-word prediction abilities, or RLHF itself, or both. RLHF is aimed at maximizing the usefulness of LLMs and for this reason, its inclusion in training is supposed to be an asset. Yet, in its presence, the LLMs' linguistic limitations outlined above are even more resounding, as they emerge despite targeted human mitigation efforts.

## Conclusions

To conclude, the present work investigated whether scaling in terms of number of parameters bridges the gap between LLM and human performance in the context of a grammaticality judgment task featuring ubiquitous properties of language: anaphoric reference, sentence embedding, comparatives and negative polarity constructions. The results showed that ChatGPT-4, the best performing model, indeed outperforms humans for accuracy in one experimental condition. While this evidence indicates that scaling matters, ChatGPT-4 does not perform better than humans in the ungrammatical condition and its instability in responses does not favor convergence towards accuracy upon repeated exposure to the same prompts.

For LLMs to be able to act as theories of natural language, their linguistic behavior should be comparable to that of humans at least at a descriptive level [2, cf. 80]. Differences in the performance of humans vs. LLMs persist, notwithstanding the fact that (i) the amount of data on which LLMs are trained vastly exceeds what humans are able to experience in a lifespan [81]; (ii) data are imbued with explicit human annotations as per their grammaticality status in models but not in humans; and (iii) the task we used was available online before the testing took place, making it possible that the tested LLMs had experience with the stimuli, since LLMs are trained on thousands of scientific papers [82].

Overall, the failure of LLMs to consistently tell apart grammatical from ungrammatical language without deviations in the judgments casts doubt on the human-likeness of their linguistic abilities. Scaling indeed matters, and different methodologies must be compared for a fuller appreciation of the models' generalization capabilities. At present, the observed differences between the language abilities of LLMs and humans seem to amount to differences of *kind*, not *scale*, that have deep roots in the process of language learning in silico vs. in vivo respectively. As such, further increments in the LLM training data are likely to mitigate these differences and mismatches, but unlikely to fully fix them.

## Data Availability Statement

All data relevant to the present study have been deposited in the OSF (https://osf.io/m6yda/?view_only=135e7be5a131458e89acc05c74247b09).